\documentclass[runningheads, a4paper]{llncs}
\usepackage{array}
\usepackage{booktabs}

\usepackage[utf8]{inputenc}
\usepackage[T1]{fontenc}
\usepackage{xcolor}

\usepackage{amsmath}
\usepackage{amsfonts}
\usepackage{amssymb}
  
\usepackage{geometry}
\usepackage{multirow}
\usepackage{graphicx}
\graphicspath{{images/}}
\def\captionformating{}

\makeatletter
\def\paragraphmark#1{ --}
\long\def\paragraph{\@startsection {paragraph}{4}{\z@ }{-12\p@ \@plus -4\p@\@minus -4\p@ }{-0.5em \@plus -0.22em \@minus -0.1em}{\normalfont \normalsize \bfseries }}
\makeatother
\begin{document}
\title{Case-based reasoning for rare events prediction on strategic sites}
\titlerunning{Case-based reasoning for rare events prediction}
\author{
Vincent Vidal \and
Marie-Caroline Corbineau \and
Tugdual Ceillier}
\authorrunning{V. Vidal et al.}

\institute{Preligens (ex-Earthcube), Paris, France \\ \url{www.preligens.com}, [name].[surname]@preligens.com}
\maketitle              % typeset the header of the contribution
\begin{abstract}
Satellite imagery is now widely used in the defense sector for monitoring locations of interest. Although the increasing amount of data enables pattern identification and therefore prediction, carrying this task manually is hardly feasible. 
We hereby propose a cased-based reasoning approach for automatic prediction of rare events on strategic sites. This method allows direct incorporation of expert knowledge, and is adapted to irregular time series and small-size datasets. Experiments are carried out on two use-cases using real satellite images: the prediction of submarines arrivals and departures from a naval base, and the forecasting of imminent rocket launches on two space bases.
The proposed method significantly outperforms a random selection of reference cases on these challenging applications, showing its strong potential.

 % Remote sensing
\end{abstract}
\section{Introduction}

In the defense sector, remote sensing, and particularly satellite imagery, is used extensively to detect events on strategic locations.
In addition, with the strong increase in the number of commercial satellite images, it has become possible to monitor the activity on specific sites over a long period of time, paving the way for identifying causal patterns on these sites. Such patterns can then be used to predict in advance events that are likely to happen on these locations.

\smallskip

Predictive analysis is a broad field, ranging from forecasting the future values of a series of observations, to detecting events before they actually occur.
This topic is very much studied in medicine~\cite{beyene2018survey,shinde2020forecasting}, ecology~\cite{dietze2017ecological} or finance~\cite{sezer2020financial}, but very little in remote sensing.
Indeed, the majority of prediction methods require to measure data at regular time intervals and to have a large number of past measurements, which is not really possible in remote sensing: depending on the satellite coverage and weather conditions, for a given location, it is very likely that, for a large number of days, no image is available.

\smallskip

In addition, it is often necessary to consider a large number of factors, sometimes on different geographical locations, to understands key patterns and be able to predict certain events.
This large increase in the number of features required for prediction makes it particularly difficult for a human operator to forecast events. Hence, the development of automatic algorithms is particularly relevant to predict defense-related events based on satellite imagery.

\subsection{Problem setting and contributions}

The problem studied here is the detection of specific events from short and irregular series of observations derived from commercial satellite images. The monitored zones are divided in subareas based on their purposes, for instance administrative areas, road check points, railroads, etc. Times series are then created from the number of objects, such as vehicles or boats, per subarea and per date.

\smallskip

We propose a method for event prediction, using a case-based reasoning approach applied to temporal fragments of satellite image series. This method was designed such that expert insight about the use-case can be directly incorporated, thus facilitating interpretation and pertinence of the results.
If the method is meant to be generic, we have focused here on two particular applications: the prediction of the arrival and departure of submarines on a naval base, and the prediction of upcoming rocket launches on launch bases. The proposed cased-based reasoning approach has been applied on real data for each use-case, and compares very favorably with a naive approach.
From our knowledge, this is the first time that such a prediction method is developed for these applications. We shall stress the fact that very few data were available, making this problem extremely challenging. These results are therefore very promising, and motivate future works on larger datasets to test the method's ability to generalize. 

\smallskip

This article is organized as follows: in Section~\ref{sec:related_work} we introduce the two families of methods that are available in the literature for predictive analysis, in Section~\ref{sec:proposed} we describe in details the proposed case-based reasoning approach, while Sections~\ref{sec:submarines} and \ref{sec:rockets} are dedicated to our experiments on real data, including discussion of the results ; finally, some conclusions are drawn in Section~\ref{sec:conclu}.

%%%%%%%%%%%%%%%%%%%%%%%%%%%%%%%%%%%%%%%%%%%%%%%%%%%%%%%%%%%%%%%%%%%%%
%%%%%%%%%%%%%%%%%%%%%%%%%%%%%%%%%%%%%%%%%%%%%%%%%%%%%%%%%%%%%%%%%%%%%

\section{Related work}\label{sec:related_work}

The problem that we are addressing is located at the junction of several fields, namely event classification, anomaly or rare event detection, and predictive analysis.
We can distinguish two main classes of methods.
On one hand, we have the methods that see the measured data as a temporal
sequence of values which can be modeled in order to predict its values in the
future.
On the other hand, we have methods that take a step back from the temporal
vision of things, cutting the sequence into fragments and treating these
fragments independently of their temporal position, which allows to easily
reduce the study to a regression or a classification problem.

\subsection{Parameterized models}

The first strategy used in predictive analysis seeks to directly model the data
set in a global way.  It assumes that the dataset has a sufficiently regular
behavior to be modeled.  If the model is manually selected to match the expected
behavior, its parameters are automatically estimated to best explain the
observations.

\smallskip

%%%%%%%%%%%% %%%% %  %             Regression             %  % %%%% %%%%%%%%%%%%
Among the many methods using this approach, we can mention the regression methods, where the data are modeled by a parametric function, whose parameters are estimated to fit the observations~\cite{zubakin2015improvement}.
  %
  %% Among the many regression methods used, we can mention the affine regression,
  %% using an affine $f$ function:
  %% \begin{equation}
  %%   f_{\omega,b}(t) = t \omega + b,
  %% \end{equation}
  %% where the parameters $\omega,b\in\mathbb R^d$, $d$ beeing the dimension of the
  %% observed data, are most often estimated by a least squares method. Note that
  %% this can easily be extended to piecewise affine models, where the sequence of
  %% values is separated into several parts, each modeled by an affine model.
  %
These regression methods are used a lot in the medical field where the studied processes (cardiac rhythm, respiratory cycle, etc) are the object of very advanced models, which only need to be parameterized to better fit each patient.
Gaussian processes are also widely used for predictive analysis~\cite{richardson2017gaussian}. They
  provide a probabilistic framework to interpolate and extrapolate temporal
  sequences while providing additional variance information with the predicted
  values.
  %% These methods are
  %% particularly interesting because of the choice of the kernel, allowing to
  %% introduce different behaviors, like periodicity or specifying different
  %% correlation scales between values.
However, these methods may not be relevant when feature space is of high dimension and when the number of data samples is low.

\smallskip

More recent approaches use deep learning, in particular the long short-term memory recurrent neural networks~\cite{sezer2020financial}. The latters process data sequentially, reusing some of the network's outputs at the next time step to retain part of the information and thus make it possible to take into account a certain temporal dependence of the observed values.
Finally, many models formulate the problem recursively, making each value depend on the previous values. We can thus mention ARIMA (Autoregressive Integrated Moving Average), which assumes a linear dependence involving Gaussian noise terms that is still used today~\cite{siami2018forecasting}, or the Grey model, which is based on the ARIMA model but expressed as a differential equation.
However, these methods require regular data and can hardly take into account missing data without serious modifications of the model.
  
%% % Gaussienne

\subsection{Sample-based approaches}

The second approach consists in separating from the notion of temporality by extracting from the studied time series different temporal segments.
Each fragment is treated as a simple point in a high dimensional space and the problem is reduced to a much more studied problem of regression or classification.
The objective is then to infer from a given segment if an event is likely to happen, rather than finding the next point of the segment.
Various methods are used in this context, for instance
support vector machines, which can be associated with the kernel trick for non-linear problems~\cite{fathima2012predictive}.
Other methods are based on fuzzy regression~\cite{azadeh2010integrated}, taking into account a certain imprecision on the data.
Small neural networks have also been used, taking time series as inputs and outputting the studied value of interest~\cite{pao2006comparing}.

\smallskip

Case-Based Reasoning (CBR) methods were theorized in the 1990s and have since been used for many applications such as energy demand prediction~\cite{platon2015hourly} or financial analysis~\cite{corchado2001hybrid}.
This approach is based on using information from a case that is similar to the
present one and that has already been solved in the past. This is in contrast to
the usual strategy of deducing a set of rules from observations to explain the
general behavior of the data.
CBR methods require inputs from the user regarding the
definition of the manipulated objects:
\begin{itemize}
\item Definition of a ``case'' or ``problem'', i.e. the variables allowing to
  identify a situation, for example for our use cases this definition could
  involve the location, the date, the number of vehicles of a certain type on
  a parking lot, etc...
\item  Defining metrics that allow computing a distance between cases to
  identify the past cases that are closest to the current one.
\end{itemize}

The majority of these methods seem to be suitable for the studied problem, provided that the irregularity of the data is taken into account. As explained in the next section, we decided to use the CBR approach.

% Peut-être un peu sec

%%%%%%%%%%%%%%%%%%%%%%%%%%%%%%%%%%%%%%%%%%%%%%%%%%%%%%%%%%%%%%%%%%%%%
%%%%%%%%%%%%%%%%%%%%%%%%%%%%%%%%%%%%%%%%%%%%%%%%%%%%%%%%%%%%%%%%%%%%%

\section{Proposed approach}\label{sec:proposed}

\subsection{Why the CBR?}
One specificity of our experiments is that available data are very scarce and irregular: as seen in Sections~\ref{sec:submarines} and \ref{sec:rockets}, between 8 and 33 images are available per use-case, and the time step between two images is not consistent. Since most of the predictive models found in literature use data samples that are regular and abundant, this issue appears to be critical.

One way to overcome data scarcity is to introduce some a priori knowledge, that is, to provide information to the model about the expected behavior or the data. This knowledge can be enforced through optimization constraints during training, or can be incorporated directly into the problem formulation, for instance by selecting a relevant subset of attributes and thus manually reducing data dimension.
Another strategy consists in generating synthetic samples to artificially increase the size of the dataset. However, this approach requires an accurate model to produce realistic data. It shall be noted that even military experts find it difficult to understand the patterns related to the two applications described in Sections~\ref{sec:submarines} and \ref{sec:rockets}. Therefore, in view of the complexity of the use-cases, and the risk to introduce a significant bias in the model, synthetic data generation was not deemed appropriate for our use-cases.
Finally, some predictive models are inherently less data greedy, this is the case, for instance, of the Grey model or of CBR, depending on the chosen prediction function. 

To carry out this project, we chose to use an approach based on CBR where the prediction function is not parametrized and therefore does not require training. Here, a case is defined by a sequence of dates, each one being associated with a list of scalar attributes, where each attribute is a number of a certain type of objects in a given subarea on the studied location. One advantage of this method is that expert insight can be incorporated into the model by appropriately selecting the object classes and subareas. 

\subsection{Data interpolation and cross-validation}\label{sec:cross-val}

As mentioned previously, one advantage of the method is that it requires few data to be applied. Nonetheless, to compare two time series we still need them to be regular, i.e. the time interval between two dates must be the same. This condition is not satisfied in our raw data since each date corresponds to a commercial satellite image and this type of imagery is not available with a regular time step. To address this issue we apply a linear interpolation on every time serie. It can be noted that, due to this interpolation, attributes or number of observables can take non integer values for interpolated dates.

\smallskip

\begin{figure}[bt]
    \centering
    \includegraphics[width=12cm]{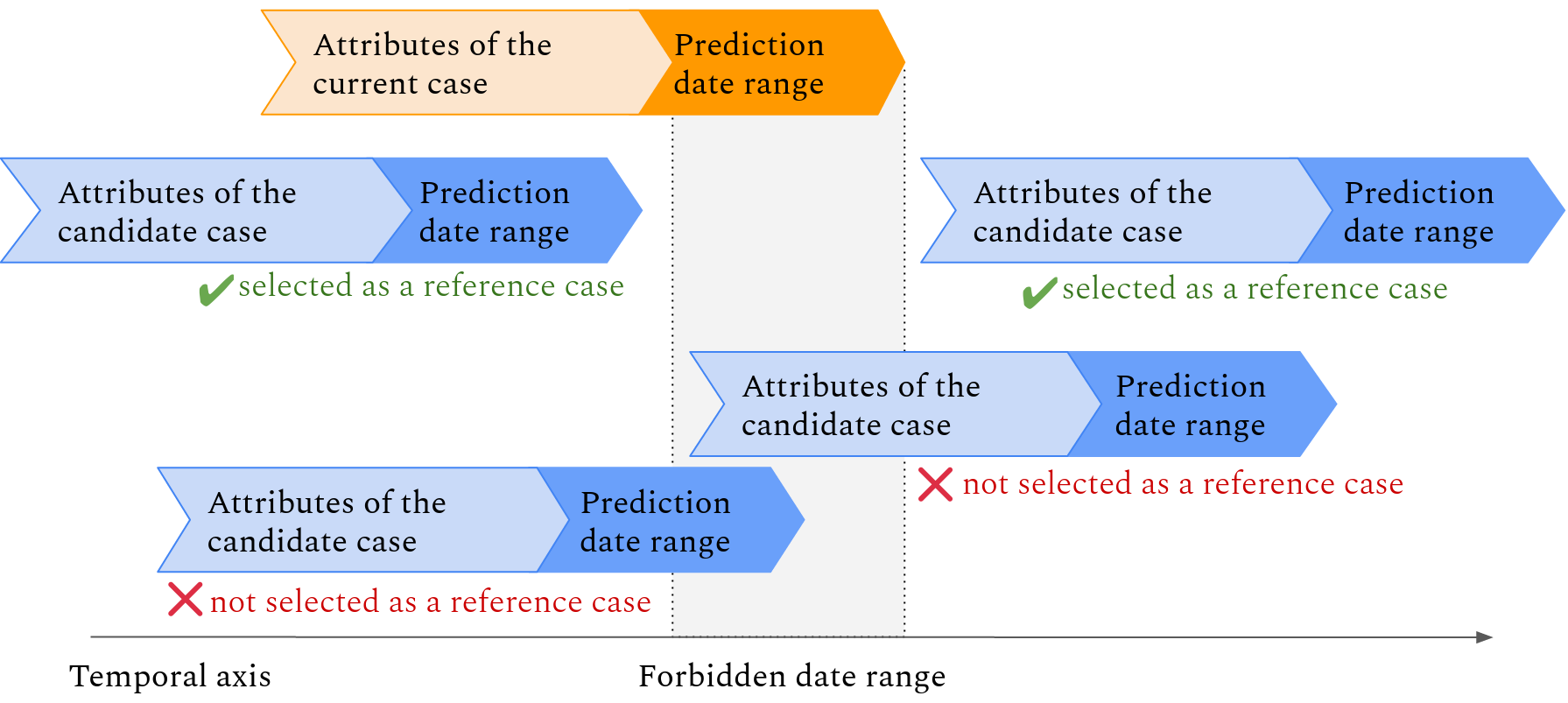}
    \caption{\label{fig:isolation_crit}%
    \captionformating%
    Illustration of the overlapping criterion for selecting reference cases.}
\end{figure}

In CBR, past cases are used to produce a prediction for the current cases. Hence, if we were to follow the chronological order, early cases would have very few reference cases to make use of and, consequently, there predictions would be less accurate. This is especially problematic here because we have very small datasets. Therefore, to ensure that our experiments are representative enough, we decided to make use of both past and future cases to generate predictions. 
To perform this cross validation in a relevant way, we set up an overlapping criterion. More precisely, as illustrated in Figure~\ref{fig:isolation_crit}, this criterion makes sure that past and future reference cases are far enough from the current case such that the prediction is not directly biased by them. 

\subsection{Detailed steps and adaptation of CBR}

\begin{figure}[bt]
    \centering
    \includegraphics[width=12.5cm]{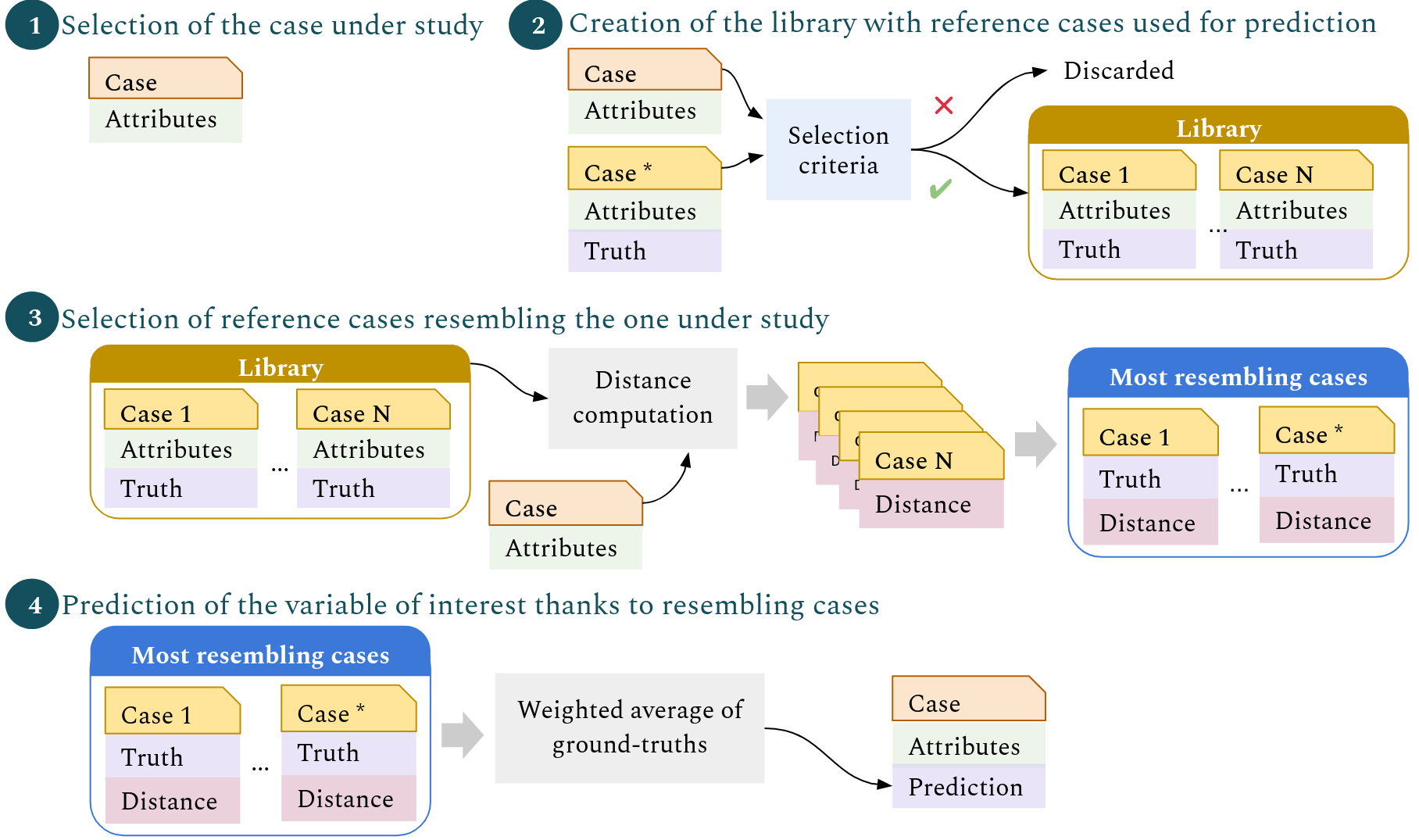}
    \caption{\label{fig:schema-method}%
    \captionformating%
    Steps involved in the proposed CBR method.}
\end{figure}

As illustrated in Figure~\ref{fig:schema-method}, the proposed approach follows four steps:
\begin{enumerate}
    \item We first select the case to be studied, that is to say a small time series where each date is associated with attributes. The list and definitions of attributes will be detailed for each experiment in Sections~\ref{sec:submarines} and \ref{sec:rockets}. The size of the time series, i.e. the number of days it is made of, is a hyperparameter of the method.
    \item For each possible case, past of future, we check if it meets the selection criteria mentioned in Section~\ref{sec:cross-val}, in particular the isolation criterion ensuring that future cases are not too close to the current case. The valid reference cases constitute the case library used for prediction, while the others are discarded.
    \item Then, the Euclidean distance between each reference case and the case under study is computed. The $K$ nearest neighbors are selected as being the most resembling cases, $K$ being set by the user.
    \item Finally, the prediction $y$ for the case under study is obtained by computing the weighted average of the ground-truths of the $K$ nearest reference cases:
    \begin{equation}\label{eq:pred}
        y = \dfrac{\sum_{i=1}^K p_i\exp\left({-\frac{d_i^2}{2\sigma^2}}\right)}{\sum_{i=1}^K \exp\left({-\frac{d_i^2}{2\sigma^2}}\right)},
    \end{equation}
    where $p_i$ is the ground-truth of the $i$th reference case and $\sigma$ is manually set to a percentage of the data standard deviation ($20\%$ in practice). Hence, the closer a reference case is to the one under study, the more weight it will have in the prediction. As seen in Eq.~(\ref{eq:pred}), weights are obtained thanks to a Gaussian kernel, so that the influence of cases that are farther away decreases faster. On the choice of $\sigma$: the larger it is, the more $y$ will tend to a simple average, and on the contrary, if $\sigma$ is small then only the $d_i$ corresponding to the closest cases will be taken into account. 
    
\end{enumerate}

In the next sections, the proposed CBR approach is applied on two different use-cases.

%%%%%%%%%%%%%%%%%%%%%%%%%%%%%%%%%%%%%%%%%%%%%%%%%%%%%%%%%%%%%%%%%%%%%
%%%%%%%%%%%%%%%%%%%%%%%%%%%%%%%%%%%%%%%%%%%%%%%%%%%%%%%%%%%%%%%%%%%%%

\section{Predicting submarines arrivals and departures}\label{sec:submarines}
The first use-case addresses the prediction of imminent arrivals and departures of submarines on a naval base, which is referred to as naval-base-1.
\paragraph{Data} For this experiment, we selected commercial satellite images based on two constraints: 1/ these images had to contain very few clouds, such that submarines could be identified, and 2/ they needed to be close to each other on the temporal axis, since the proposed predictive method makes use of time series. We acquired all available images of naval-base-1 satisfying these criteria, leading to a total of 33 images, with 3 images in June 2018, 3 in September 2019 and 27 in 2020.
\paragraph{Feature extraction} The area of interest in naval-base-1 is divided into two zones based on the analysis of experts at Preligens. For each zone, we manually determined the number of vessels belonging to the following 9 classes for every date: relevant submarine classes (Delta III, Oscar II, Borei and Akula), warships, support ships (including tugboats), barge-mounted cranes, speedboats and civilian boats. In addition, zone 2 also exhibited a parking lot, so we counted the number of vehicles on this area for each image.  
\paragraph{Formulation of the prediction task} Given the nature of the studied naval base, the following two variables were defined as prediction targets:
\begin{itemize}
    \item the arrival of at least one submarine of all classes in the next 4 days,
    \item the departure of at least one submarine of all classes in the next 5 days.
\end{itemize}
It shall be noted that we are not interested in predicting the number of submarines, since this variable could stay constant even if there have been multiple departures and arrivals. The separation between the two tasks above enriches the study, and is made possible thanks to the precise identification of each submarine's class. 
In addition, we are not trying to estimate the number of arrivals and departures, but rather to predict if at least one such event is likely to happen. This choice is motivated by the fact that only a small amount of data is available, which makes this study challenging. Also, we do not have one image for each day in the studied period, so the ground-truths we used for the predicted events are approximations. Indeed, in the days following the prediction date, there could be an arrival or departure of submarine on a date for which no image is available.
\paragraph{Hyperparameters}
The proposed CBR approach includes several hyperparameters that are chosen manually. Several configurations were tried and we kept the one leading to the best predictions. Regarding the prediction support, which is the number of previous days used to estimate the likelihood of the studied events, we used time series of 9 days preceding the prediction date. For both arrivals and departures we used a total of 5 nearest neighbours to make the prediction. Finally, each attribute is defined by a pair (class, area), where classes could also be merged categories. From all possible attributes we kept only the most relevant ones: the number of elements for each one of the four submarine classes in zone 1, the number of warships in zone 1 and the number of boats (all categories are mixed) in zone 2. Regarding the arrival of submarines, we included an additional attribute, which is the number of vehicles in zone 2.  
\paragraph{Results and discussion}
To complement the study, we compared the proposed CBR method with a random approach, where the neighbors that are selected in the library to infer the prediction are chosen randomly before applying Eq.~\eqref{eq:pred}. For this comparison, we take the average of 10 random draws. Results for both the proposed approach and the random method are presented in Table~\ref{tab:naval-base-1-res}. The proposed CBR model correctly predicts all actual arrivals and departures in zone 1, and leads to only two false alarms: one arrival and one departure that are expected by the model but not visible in the images.

\begin{table}[!htbp]
\centering
\caption{\label{tab:naval-base-1-res}Prediction of submarines activity on naval-base-1. True positive: an event is correctly predicted. True negative: the absence of event is correctly predicted.}
\begin{tabular}{rr*5c}
\toprule
&~~&  \multicolumn{2}{c}{Submarine arrivals (4 days)} & & \multicolumn{2}{c}{Submarine departures (5 days)}\\
\midrule
 & ~~   & True positives   & True negatives &~~~~& True positives   & True negatives \\
Nearest neighbors & ~~  &  \textbf{3/3} & \textbf{7/8}   &~~~~& \textbf{6/6}  & \textbf{7/8} \\
Random & ~~   & 2.5/3 & 6.5/8   &~~~~& 3.7/6  & 4.7/8 \\
\bottomrule
\end{tabular}
\end{table}

In addition, the CBR approach compares well with the random draws, which suggests that the model positively identifies "patterns" in the data that help producing correct predictions.
To check whether these patterns are causal or just lucky correlations, we carry out an ablation study. In other words, we gradually remove attributes and study how the prediction is affected. For the submarines arrivals, this process suggests that most of the information is encased in the number of vehicles in zone 2. From an operational point of view, zones 1 and 2 do not serve the same purpose, thus the pattern used by the model to predict arrivals is likely to be a lucky correlation in this case. For departures however, the information seems to be diluted in the different attributes that we selected. Therefore, the associated patterns might be more relevant and would benefit from a more in-depth study if more data is available.
\\
It is worth noticing that other configurations were studied, some including for instance the number of barge-mounted cranes in the list of attributes. However, these additional attributes, despite being relevant from an operational point of view, did not raise the performance. This might be explained by their small representation in the data and the restricted number of images that were available.

\section{Predicting rocket launches}\label{sec:rockets}
This second use-case aims at predicting imminent rocket launches on two different space launch bases, namely space-base-1 and space-base-2.
\paragraph{Data} We retrieved online the official dates of past launches on the two locations, and gathered all available sequences of commercial satellite images close to these dates. We were able to acquire 8 images for space-base-1 and 14 images for space-base-2, that were associated to 3 launch dates for both locations.
\paragraph{Feature extraction} The definition of relevant areas in the two launch bases was provided by a group of experts at Preligens. Then, using Preligens' vehicle detectors, we were able to automatically get the number of vehicles in each subarea for every image. In addition, for space-base-1 we manually annotated some elements that might be of interest from an operational point of view.
%19-08-2021: the number of wagons at the end of rail roads leading to the launch pads, since this element 

%
\paragraph{Formulation of the prediction task} The goal of the CBR model for this application is to predict an imminent rocket launch on the current date or in the next 4 or 5 days on space-base-1 and space-base-2, respectively.
Here, we do not try to predict the launch pads on which the event will happen. There are two reasons for this: first, in view of the small amount of data we would have a very low number of events per launch pad, and second, this information is not always publicly available. In addition, we are not trying to predict a precise launch date since the database would not be sufficient for a problem of that complexity.
\paragraph{Hyperparameters} The support size is 1 for this application: we use the image of only one date to infer whether a launch is likely to happen in the next days. Increasing the support size would lead to very few valid cases in view of our overlapping criterion illustrated in the Figure~\ref{fig:isolation_crit}. For every prediction we select 3 nearest neighbors in the case library. We select the attributes that are most relevant to the application and that lead to the best results:
\begin{itemize}
    \item for space-base-1 we use 6 attributes, including the number of vehicles in 5 different areas (administration, preparation, launch pads, maintenance, road check-point)
    %19-08-2021: and the aforementioned number of wagons near the launch pads ;
    \item for space-base-2 we use the number of vehicles in 3 subareas (administration, preparation and launch areas).
\end{itemize}
For space-base-1, the number of vehicles per subarea is in average much higher than the last attribute.
%19-08-2021: number of wagons. 
Therefore, in order to avoid an imbalance in the computation of the Euclidean distance, we artificially divide the number of vehicles on space-base-1 by 10. This can be viewed as a type of normalization.
\\
It shall be noted that the output of the model is a number between 0 and 1. For the prediction of submarines arrivals and departures, the output was numerically sufficiently close to either 0 or 1 so that we did not need to tune a threshold to decide whether the final prediction was 0 or 1. Here, it is different, predictions can take intermediary values. Hence, we chose to take a threshold equal to 0.1 to avoid missing launches: if the launch score is above 0.1, then we set the prediction to 1.
\paragraph{Results and discussion} Results are presented in Table~\ref{tab:space-base-res}, similarly to the previous use-case, we include a comparison with a model using random selection of neighbors.
The proposed approach leads to one false alarm on space-base-1, while all 14 predictions are correct on space-based-2. The CBR model compares favorably with the random approach: for space-base-1 the latter leads to more than 2 false alarms in average, and for space-base-2 it misses more than half the launches. This suggests that the model identifies relevant patterns.

\begin{table}[!htbp]
\centering
\caption{Results for the prediction of imminent rocket launches. True positive: an event is correctly predicted. True negative: the absence of event is correctly predicted.}\label{tab:space-base-res}
\begin{tabular}{rr*5c}
\toprule
&~~&  \multicolumn{2}{c}{space-base-1 (4 days)} & & \multicolumn{2}{c}{space-base-2 (5 days)}\\
\midrule
 & ~~   & True positives   & True negatives &~~~~& True positives   & True negatives \\
Nearest neighbors & ~~  &  \textbf{5/5} & \textbf{2/3}   &~~~~& \textbf{6/6}  & \textbf{8/8} \\
Random & ~~   & 4.0/5 & 0.7/3   &~~~~& 2.9/6  & 7.0/8 \\
\bottomrule
\end{tabular}
\end{table}

\begin{figure}[bt]
    \centering
    \includegraphics[width=12cm]{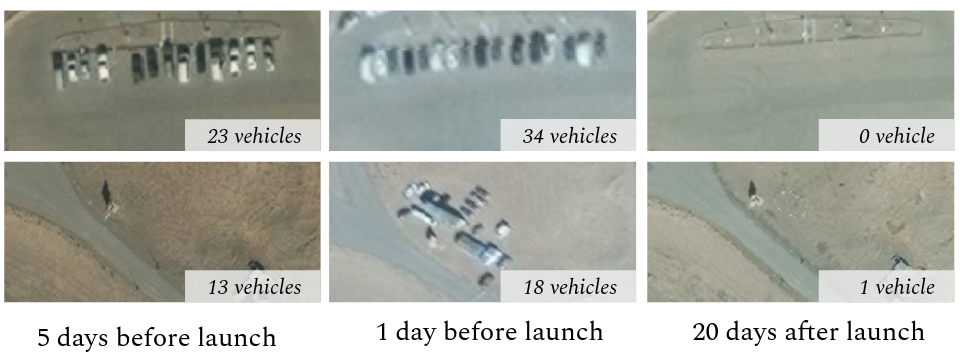}
    \caption{\label{fig:space-base-2}%
    \captionformating%
    Number of vehicles before and after a launch on space-base-2. Close-ups of preparation (first row) and launch (second row) subareas. Indicated numbers of vehicles are taken on the full subareas so not all vehicles are visible.}
\end{figure}

Regarding the first location, the ablation study clearly demonstrated that mainly one attribute, which is relevant from an operational point of view, is the key pattern.
%19-08-2021: the key pattern is the number of wagons on the rails near the launch pads, which is relevant from an operational point of view.
%
For space-based-2, the ablation studies shows that the attribute that provides less information is the number of vehicles in the administrative areas: when considering only this attribute, the model makes 3 mistakes compared to none when the other dimensions are used individually. This could be explained by a low number of vehicles on average in administrative areas (usually below 5), when other areas feature larger variations, from one to several dozens of vehicles depending of the date.
Fig.~\ref{fig:space-base-2} illustrates this phenomenon: we can see that the number of vehicles increases dramatically ahead of launches.

Here, the patterns that seem to be exploited by the model can be inferred by a human as well: we are dealing with few images and the evolution of the observables of interest
%19-08-2021: vehicles or wagons 
the days prior to a launch is quite obvious. Hence, we do not claim that this model has discovered unknown patterns, but rather that the behavior of the proposed method is reassuring and in accordance with operational consideration, showing its potential for other, more complex, applications.

%%%%%%%%%%%%%%%%%%%%%%%%%%%%%%%%%%%%%%%%%%%%%%%%%%%%%%%%%%%%%%%%%%%%%
%%%%%%%%%%%%%%%%%%%%%%%%%%%%%%%%%%%%%%%%%%%%%%%%%%%%%%%%%%%%%%%%%%%%%

\section{Conclusion}\label{sec:conclu}

In addition to being applicable to very irregular time series, the proposed approach has the benefit of easing the interpretation of the prediction, as the selection of the nearest neighbors and their distance from the study case allows a better understanding of the patterns present in the data and related to the variable of interest to be predicted.
If the results of the method on the two use cases presented are very promising, it is important to note that the small amount of data used does not allow to fully validate the method. To do so, it would be necessary to perform additional tests on larger data sets.

Moreover, to obtain a prediction value from the neighbors, we use here a simple weighted average. If the large size and the small number of data prevent us from using advanced methods (such as deep neural networks), other methods (such as SVM) could be used in future works to bring more expressivity to the model.
Finally, it could be interesting to study the use of other distance measures less sensitive to missing data (using for example probability distributions) or which could take into account temporal deformations of the studied patterns (in the line of the dynamic time warping).

\subsection*{Acknowledgment}

This work is supported by the French "Direction Générale de l'Armement" through the research project RAPID S-Cube in collaboration with Preligens. 
\bibliographystyle{splncs04}
\bibliography{references}

\end{document}